\documentclass[runningheads]{llncs}
\usepackage{url}
 
\usepackage{makecell}
\usepackage{array}
\usepackage{booktabs}
\usepackage{graphicx}
\usepackage{amsmath}
\usepackage{multirow}
\usepackage[ruled]{algorithm2e}

\newcommand{\ie}{\textit{i.e.}}
\newcommand{\eg}{\textit{e.g.}}
\newcommand\blfootnote[1]{%
  \begingroup
  \renewcommand\thefootnote{}\footnote{#1}%
  \addtocounter{footnote}{-1}%
  \endgroup
}

\begin{document}
\title{Colorectal Polyp Classification from White-light Colonoscopy Images via Domain Alignment}
\titlerunning{Colorectal Polyp Classification from White-light Colonoscopy Images}

\author{
Qin Wang$^{1,2\dagger}$\and 
Hui Che$^{1,2\dagger}$\and  
Weizhen Ding$^{1,2}$\and    
Li Xiang$^{3}$\and          
Guanbin Li$^{2,4}$\and      
Zhen Li$^{1,2*}$\and        
Shuguang Cui$^{1,2}$\\      
\email{\{qinwang1@link., lizhen@\}cuhk.edu.cn,huiche.stu@gmail.com}
}
\institute{
$^1$The Chinese University of Hong Kong(Shenzhen)\\
$^2$Shenzhen Research Institute of Big Data\\
$^3$Longgang District People's Hospital of Shenzhen\\
$^4$Sun Yat-sen University
}


%
\maketitle              
\begin{abstract}
Differentiation of colorectal polyps is an important clinical examination. A computer-aided diagnosis system is required to assist accurate diagnosis from colonoscopy images. Most previous studies attempt to develop models for polyp differentiation using Narrow-Band Imaging (NBI) or other enhanced images. 
However, the wide range of these models' applications for clinical work has been limited by the lagging of imaging techniques.
%
Thus, we propose a novel framework based on a teacher-student architecture for the accurate colorectal polyp classification (CPC) through directly using white-light (WL) colonoscopy images in the examination.
In practice, during training, the auxiliary NBI images are utilized to train a teacher network and guide the student network to acquire richer feature representation from WL images. 
The feature transfer is realized by domain alignment and contrastive learning. Eventually the final student network has the ability to extract aligned features from only WL images to facilitate the CPC task. 
Besides, we release the first public-available paired CPC dataset containing WL-NBI pairs for the alignment training. 
Quantitative and qualitative evaluation indicates that the proposed method outperforms the previous methods in CPC, improving the accuracy by {\textbf{5.6\%}} with very fast speed.

\end{abstract}
\blfootnote{$*$ Corresponding author. $^\dagger$ Equal first authorship.}
\section{Introduction}
Colorectal cancer (CRC) is one of the most common malignancies with a high mortality rate around the world~\cite{chen2018accurate}. 
Colorectal polyps are recognized as indicators of CRC, and they 
%
are roughly classified into two categories: hyperplastic and adenomatous~\cite{fonolla2020cnn}. 
Hyperplastic polyps are benign while adenomatous polyps have a high possibility of malignant transformation. 
Considering only the latter ones are required for surgical resection, precise differentiation is important to decrease unnecessary resection and unsuitable treatment. 
Colonoscopy is the preferred detection and diagnostic tool for colorectal polyps. 
However, due to varying illumination conditions, similar tissue representation, and occlusion, it is usually difficult to discriminate between benign and pre-cancerous polyps by conventional white-light (WL) observation, even for well-experienced endoscopists~\cite{komeda2017computer}. 
%
Therefore, an accurate and objective computer-aided classification system is demanded to assist clinical work.

Recent studies have achieved promising performance in colorectal polyp classification (CPC) by employing deep learning-based methods. 
Most works prefer to use datasets containing Narrow-Band Imaging (NBI) or Blue Light Imaging (BLI) images, owing to the enhanced visibility and superrior performance~\cite{rondonotti2019blue}.  
For example, Usami {\it et al.}~\cite{usami2020colorectal} proposed to distinguish benign/malignant polyps using WL, dye, and NBI images. 
%
In~\cite{fonolla2020cnn}, authors achieved the highest accuracy of 95\% by combining WL, BLI, and Linked Color Imaging (LCI) modalities. 
\begin{figure}[!t]
    \centering
    \includegraphics[width=0.7\textwidth]{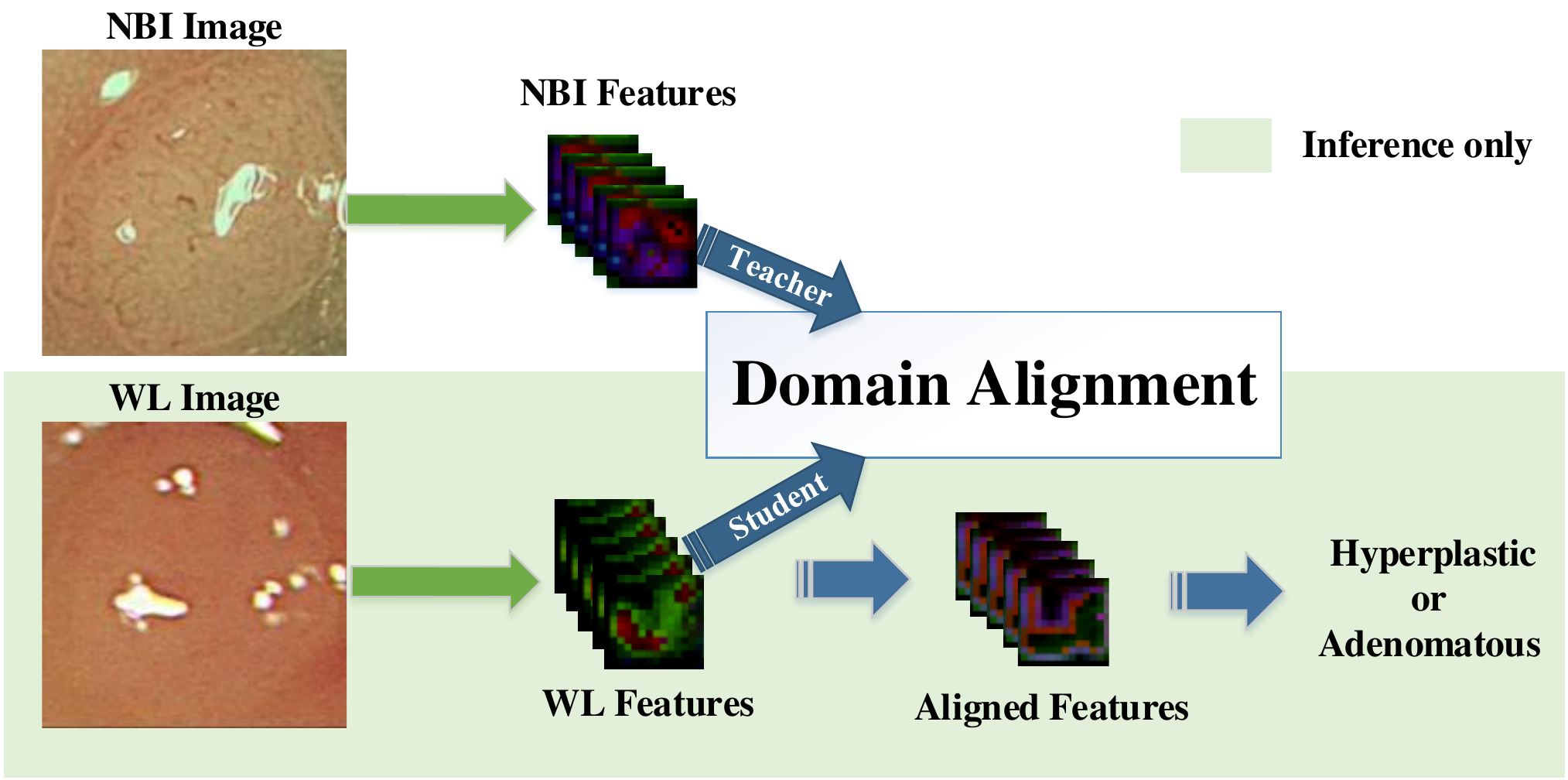}
    \caption{The proposed teacher-student approach for polyp classification only utilizing WL images during inference. 
    To improve the WL image based polyp differentiation accuracy, we adopt domain alignment to shift the distribution from WL features to NBI features during training, with the assistance of corresponding paired NBI images. 
    %
    }
    \label{fig:trailer}
\end{figure}

Nevertheless, the widely used colonoscopy devices only have WL and NBI modes.
%
%
Moreover, the acquisition of those advanced images is required to switch manually when polyps have been detected, while it usually suffers from missing detection in real clinic scenarios.  
%
Thus, the colorectal polyp detection using only WL endoscopy images is important but has not drawn sufficient attentions. 
Recently, Yang {\it et al.}~\cite{yang2020automated} reported the classification results using WL images with the accuracy 79.5\%. 
As shown, there is a large gap for the classification accuracy between using WL endoscopy images and enhanced images. 
In this paper, we propose a novel framework as illustrated in Fig.~\ref{fig:trailer} to facilitate the CPC task from WL colonoscopy images. To enhance low representative WL features, we adopt domain alignment to minimize the distance between WL and NBI feature distributions. Better feature representation in NBI images is transferred to the student network through domain alignment using adversarial learning and contrastive learning. 
Our main contributions are summarized in three-fold: (1) Through experiments, we prove that the CPC accuracy using WL images is nearly 10\% lower than that of using NBI images (88.9\%) as input. Based on this observation, we define a new scheme that exploits NBI features to improve the WL-based classification results. (2) We propose a teacher-student model with GAN-based domain alignment and contrastive learning strategies to improve CPC. (3) We further release the first public-available polyp classification dataset named CPC-Paired, including WL-NBI image pairs. Our method achieves state-of-the-art performance (\ie, $\sim 6\%$ improvement).

\section{Related Work}

\subsubsection{Domain Alignment (DA).} 
DA methods aim to align feature distributions between the source and target domains. Deep CORAL~\cite{sun2016deep} defines a loss function to constrain the distance between the source and target domains in deep layer activations. In~\cite{morerio2017minimal}, correlation alignment is connected with entropy minimization to provide a solid performance. 
The above methods are applicable when target labels cannot be accessed. 
Other methods turn to minimize the difference between the source and target distributions in a shared feature space. 
The joint maximum mean discrepancy (JMMD)~\cite{long2017deep} is introduced to learn a transfer network by aligning the joint distributions of the network activations in domain-specific layers. 
Adversarial learning is adopted in domain adaptation to learn representations that the discriminator cannot distinguish between domains~\cite{tzeng2017adversarial,zhang2018collaborative}. 
In this paper, we adopt this concept to align the features in different domains.



\section{Method}

\subsection{Adversarial Learning for Domain Alignment}
\label{sec:gan}
\begin{figure}[!t]
    \centering
    \includegraphics[width=0.95\textwidth]{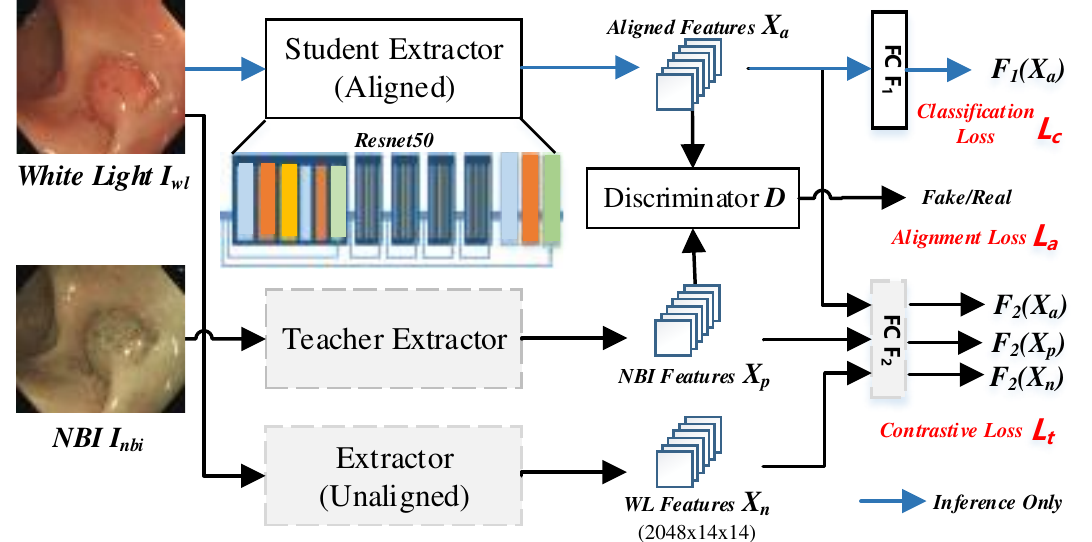}
    \caption{The overview pipeline of the proposed method. First, we pretrain teacher and unaligned extractors by utilizing NBI images and WL images separately, which is shown in grey parts. 
    To align the WL features $X_a$ with more representative NBI features, an alignment loss $\mathcal{L}_a$ is designed to optimize the student extractor by introducing a discriminator for adversarial learning. 
    Particularly, the discriminator aims to distinguish WL and NBI features. 
    Sec.~\ref{sec:gan} illustrates details about alignment loss. 
    Finally, the aligned accurate features $X_a$ are fed in fully connected layer $F_1$ for polyp classification. 
    Moreover, we exploit the contrastive learning loss $\mathcal{L}_t$ to shift the aligned WL features $X_a$ much closer to NBI features $X_p$  and far from the unaligned WL features $X_n$, which is introduced in details in Sec.~\ref{sec:contrastive}.
    The blue arrow path indicates the inference phase. 
    }
    \label{fig:pipeline}
\end{figure}

Inspired by~\cite{sankaranarayanan2018generate}, we adopt generative adversarial networks (GAN) to align the WL features with NBI features. 
As shown in Fig.~\ref{fig:pipeline}, a teacher-student scheme is designed for the feature alignment. 
More specifically, we first pretrain a teacher feature extractor by only utilizing NBI images for CPC, where rich features can be extracted from NBI images to classify polyps. 
Then, we fix the teacher extractor to output NBI features $X_p$ for aligning features  $X_a$ from student extractor. 
Particularly, the student extractor aims to extract features from WL images for polyp classification. 
However, the WL features $X_a$ extracted from WL image are unsatisfactory for accurate polyp classification, rather than the features $X_p$ from NBI images. 
Hence, to improve the classification accuracy of WL images, a discriminator $D$ is introduced to align the WL features $X_a$ with the rich NBI features $X_p$. The discriminator is optimized to distinguish between aligned WL features $X_a$ and NBI features $X_p$ (\ie, NBI features are real and WL features are fake). 
As same with the GAN training manner, the discriminator $D$ and student extractor are optimized alternatively. 
Therefore, the adversarial loss $\mathcal{L}_a$ supervises the student extractor to align its output with the teacher's (\ie,  NBI features $X_p$), which is shown in Eq.~\ref{eq:adv} where $CE$ is the cross-entropy loss, $Y_{nbi}$ indicates real label and takes 1 in practice. 

\begin{equation}
\label{eq:adv}
\mathcal{L}_a  = CE(D(X_a), Y_{nbi}) = -log(D(X_a))Y_{nbi}
\end{equation}

\subsection{Contrastive Learning on CPC}
\label{sec:contrastive}
As shown in Fig.~\ref{fig:pipeline}, to further facilitate the model convergence and boost the performance, we design a novel contrastive loss $\mathcal{L}_t$ to take advantage of contrastive learning. 
 More specifically, a naive unaligned feature extractor is pretrained to extract WL features $X_n$ for CPC by only utilizing WL images as input. 
 Then, the contrastive loss $\mathcal{L}_t$ can be formulated to supervise the student extractor to generate more representative features which are more similar to NBI features $X_p$ and dissimilar to WL features $X_n$. 
 Particularly, we take NBI features $X_p$ as positive samples and unaligned WL features $X_n$ as negative samples. 
 To optimize aligned features $X_a$, the Kullback-Leibler (KL) divergence is adopted to constrain the distribution distance from $X_a$ to WL features $X_n$ (\ie, negative samples) and NBI features $X_p$ (\ie, positive samples) in high-level semantic space, which is shown in Eq.~\ref{eq:kl}. And $F_2$ is the fully connected (FC) layer to take feature maps for probability vectors generation, which is pretrained with the teacher extractor for classifying NBI images. 
 
\begin{equation}
\label{eq:kl}
KL_{margin}(F_2(X_a),F_2(X_p)) \leq KL_{margin}(F_2(X_a),F_2(X_n))
\end{equation} 
Finally, the triplet loss $\mathcal{L}_t$ is defined for contrastive learning in Eq.~\ref{eq:cont}, where $\mu$ is a hyper-parameter we set as 0.85 in practice.

\begin{equation}
\label{eq:cont}
\mathcal{L}_{t} = \max(KL\left(F_2(X_a),F_2(X_p)\right)-KL\left(F_2(X_a),F_2(X_n)\right)+\mu, 0)
\end{equation}

\subsection{Loss Function}
The overall training loss $\mathcal{L}= \mathcal{L}_c + \mathcal{L}_a + \mathcal{L}_t$ contains three parts. 
First, a conventional cross-entropy loss $\mathcal{L}_c=CE(F_1(X_a),Y)$ is applied to supervise student extractor and FC layer $F_1$ for binary classification (\ie, hyperplastic or adenomatous), where $Y$ is the ground truth.
Then, the alignment loss $\mathcal{L}_a$ and contrastive loss $\mathcal{L}_t$ are utilized to align the WL features with NBI features, which make use of GAN and contrastive learning separately. 
Three loss functions are optimized jointly with equal weights. 
The Algorithm~\ref{alg} illustrates the whole training and inference procedures.

\begin{algorithm}[t]
\label{alg}
\caption{WL Image CPC via Domain Alignment}
\LinesNumbered 
\KwIn{NBI Images $I_{nbi}$;  Paired WL Images $I_{wl}$; CPC Label $Y$; 
Test WL Images $I_s$}
\BlankLine
Pretrain $Extractor_{teacher}$ and FC $F_2$ by $I_{nbi}$ only;\\
Pretrain $Extractor_{unaligned}$ by  $I_{wl}$ only;\\
\BlankLine
//Training Phase \\
For $I_{nbi}^i, I_{wl}^i, Y^i$ in $\{I_{nbi}, I_{wl},Y\}$\\
 \Indp
 //Extract Features\\
$X_p \longleftarrow Extractor_{teacher}(I^i_{nbi})$\\
$X_a \longleftarrow Extractor_{student}(I^i_{wl})$\\
$X_n \longleftarrow Extractor_{unaligned}(I^i_{wl})$\\
\BlankLine
//Train Discriminator $D$\\
Minimize Loss $CE(D(X_p), 1)$ + $CE(D(X_a), 0)$\\
\BlankLine
//Train Student Extractor\\
Minimize CPC Loss $\mathcal{L}_c=CE(F_1(X_a), Y^i)$\\
Fix $D$ and Minimize Alignment Loss $\mathcal{L}_a=CE(D(X_a), 1)$\\
Minimize Contrastive Loss $\mathcal{L}_t \leftarrow Triplet(X_p, X_a, X_n)$\\
\Indm
End\\
\BlankLine
//Inference Phase\\
$\hat{Y} \longleftarrow F_1(Extractor_{student}(I_{s}))$\\

\textbf{Output}: CPC Prediction $\hat{Y}$
\end{algorithm}





\section{Experiments and Results}

\subsection{Implementation Details}
We implement our work by PyTorch.  
All models were trained for 500 epochs by Adam optimizer with learning rate $10^{-3}$ and weight decay $10^{-8}$ on single Nvidia V100 GPU. 
We randomly split the dataset into training and validation set by ratio $ 8:2$ for training and evaluation. 
The training batch size is $16$. 
We adopt random flipping and rotation for data augmentation. Additionally, we apply 5-fold cross-validation for all experiments, which randomly generates 5-fold train-valid settings. 

\subsection{Dataset and preprocessing}
We conduct the experiments on our CPC-Paired dataset.
The paired data means each WL image has a corresponding NBI image with the same polyp label. For each modal, a total of 307 adenomatous and 116 hyperplastic images are included. Our dataset consists of two parts: collated data from ISIT-UMR Colonoscopy Dataset~\cite{mesejo2016computer} and clinical data collected from the hospital.
ISIT-UMR Colonoscopy Dataset contains 76 short video sequences with category information. For our CPC task, we choose 21 hyperplastic lesions and 40 adenomas sequences. Each lesion in the video is recorded using both NBI and WL. We extract paired frames from videos to build an available dataset. The eventual collated data covers 102 adenomatous and 63 hyperplastic images in each modal.
In addition, we collected 258 WL-NBI image pairs from 123 patients consisting of 205 adenoma images and 53 hyperplastic polyp images.  
We further annotate the bounding box of polyps to crop the corresponding area and scale it to $448\times 448$ as input for the CPC task.
%

%
%

\subsection{Network Architecture}
In our framework, three extractors share the same backbone design. 
The backbone can be popular network architectures (\eg, VGG, ResNet50, Inception-V3).  
More specifically, each extractor is utilized to mapping the original NBI $I_{nbi}$ or WL images $I_{wl}$ to a high-level feature space with the shape $2048\times 14\times 14$ (\eg,  ResNet50 backbone). 
Finally, each extractor is followed with a single FC layer to predict the final polyp class. 
In the pretrain stage, extractors and FC layers are optimized jointly(\eg, teacher extractor and FC layer $F_2$) which will be fixed during the alignment training phase. 
The discriminator $D$ consists of two convolution layers and two fully connected layers which aims to distinguish aligned WL features $X_a$ and NBI features $X_p$.

\begin{table}[t]
\centering
\caption{The comparison between our approach and the previous best method in~\cite{yang2020automated} .  
From the comparison, we can clearly notice that our approach surpasses the previous approach with a large margin(\eg, $\sim 6\%$) among all backbones. 
` FOLD X' indicates the different cross-validation split settings. 
`Speed' indicates the inference time per image in millisecond. 
The ` Mean' averages the accuracy among all split settings, which gains 5.6\% improvement by our approach and exactly proves the superior performance of the proposed alignment method. 
}
\renewcommand\arraystretch{1.1}
\begin{tabular}{p{1.8cm}<{\centering}p{1.8cm}<{\centering}p{1.2cm}<{\centering}p{1.1cm}<{\centering}p{1.1cm}<{\centering}p{1.1cm}<{\centering}p{1.1cm}<{\centering}<{\centering}p{1.1cm}<{\centering}<{\centering}p{1.1cm}<{\centering}}
\toprule[0.5pt]
\specialrule{0em}{1pt}{0.5pt}
     & Backbone          & Speed          & FOLD1  & FOLD2  & FOLD3  & FOLD4 & FOLD5 & Mean \\  \specialrule{0em}{0.5pt}{0.5pt}     \hline  \hline \specialrule{0em}{0.5pt}{1pt}
Yang~\cite{yang2020automated} & \multirow{2}{*}{VGG}    & \multirow{2}{*}{20.62 ms}  & 78.2\% & 79.5\% & 77.3\% & 78.0\% & 77.9\% & 78.2\%   \\
Our &              &             & 79.4\%  & 81.1\%  & 78.5\%  & 79.1\%  & 80.1\%  & 79.7\% \\   \specialrule{0em}{0.5pt}{0.5pt}     \hline  \specialrule{0em}{1pt}{1pt}
Yang~\cite{yang2020automated} & \multirow{2}{*}{\begin{tabular}[c]{@{}c@{}}InceptionV3\end{tabular}} & \multirow{2}{*}{30.72 ms} & 81.1\% & 82.1\% & 80.5\% & 82.0\% & 81.3\% & 81.5\% \\
Our &             &              & 84.6\%  & 85.7\%  & 83.1\%  & 85.3\%  & 84.4\% & 84.6\% \\   \specialrule{0em}{0.5pt}{0.5pt}     \hline  \specialrule{0em}{0.5pt}{0.5pt}

Yang~\cite{yang2020automated} & \multirow{2}{*}{ResNet50} & \multirow{2}{*}{17.07 ms} & 79.5\% & 80.5\% & 78.0\% & 80.3\%   & 78.8\% & 79.7\% \\
Our &               &            & \textbf{85.9\%} & \textbf{86.1\%} & \textbf{84.2\%} & \textbf{85.8\%} & \textbf{85.2\%} & \textbf{85.3\%}\\   \specialrule{0em}{0.5pt}{0.5pt}     \hline \hline \specialrule{0em}{0.5pt}{0.5pt}
Our w/o DA & \multirow{2}{*}{ResNet50} & \multirow{2}{*}{17.07 ms} & 82.6\% & 83.3\%  & 81.1\%  & 83.7\%  & 83.3\% & 82.9\%  \\
Our w/o CL &            &               & 83.0\% & 84.1\%  & 83.5\%  & 84.4\%  & 84.0\% & 84.0\% \\   \specialrule{0em}{0.5pt}{0.5pt}      
\bottomrule[0.5pt]
\end{tabular}
\label{tab:cmp}
\end{table}

\subsection{Results}
Extensive experiments are conducted to demonstrate the superior performance of the proposed approach. 
Particularly, by 5-fold cross comparison in Tab.~\ref{tab:cmp}, we can observe that our method outperforms previous state-of-the-art approach~\cite{yang2020automated} among all folds. 
Specifically, we improve the accuracy on all backbones including VGG, Inception-V3 and  ResNet50, which proves the generality of the proposed model. 
We obtain the best performance of our approach with ResNet50 backbone  (\ie, the highest classification accuracy 85.3$\%$, and 5.6$\%$ improvement on the mean score compared to the previous method).  
The ablation study further examines the gains of each component within the proposed approach as shown in Tab.~\ref{tab:cmp}. 
`Our w/o DA' indicates removal of alignment loss and `Our w/o CL' indicates ablation of contrastive loss. 
The CPC accuracy degradation of the ablation study exactly proves the effectiveness of each proposed component. 

\begin{figure}[!t]
    \centering
    \includegraphics[width=\textwidth]{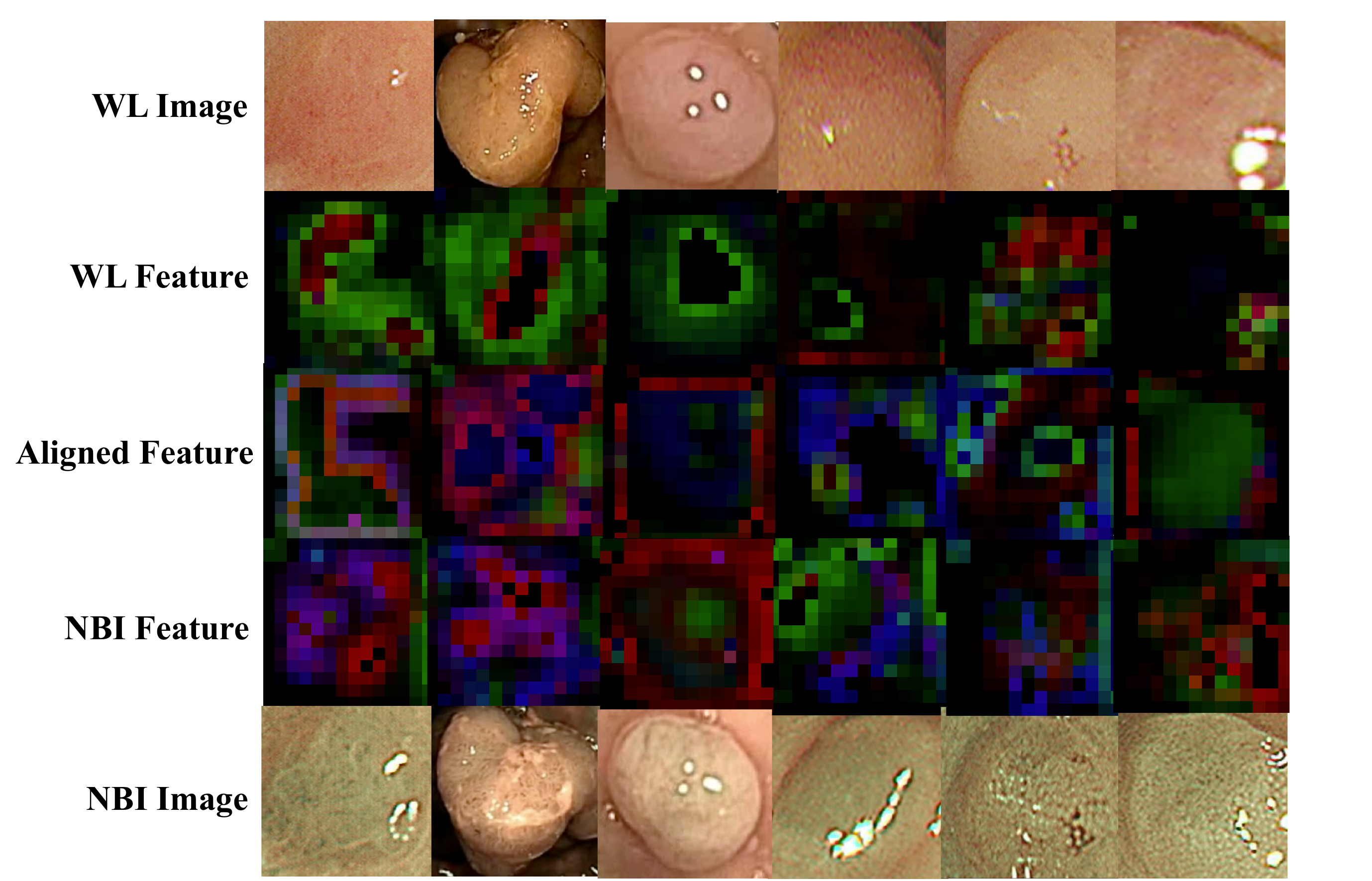}
    \caption{The visualization comparison between WL feature, aligned feature, and NBI feature. From the comparison, we can obviously notice that the aligned feature (third row) is more similar to the NBI feature (fourth row) and less similar to the WL feature (second row), which exactly prove the effectiveness of the proposed domain alignment and contrastive learning approaches for domain shifting from WL to NBI. The aligned WL feature not only contains the original WL information but also provides more essential NBI domain representation for polyp classification.}
    \label{fig:vis}
\end{figure}

The qualitative analysis is shown in Fig.~\ref{fig:vis}. 
Particularly, we extract aligned WL features $X_a$, unaligned WL features $X_n$ and NBI features $X_p$ for comparison. 
Obviously, the aligned WL features are more similar to NBI features than unaligned ones, which further demonstrates the superiority of aligned features and improvement of the proposed alignment approach. 

\section{Conclusion}

For the purpose of investigating CPC, we release a polyp classification dataset CPC-Paired. To the best of our knowledge, this is the first public-available dataset including WL-NBI image pairs for this task. To improve the CPC accuracy of white-light (WL) images, we propose a teacher-student model for shifting the feature domain of WL images to NBI images which will be more representative for the CPC. 
Particularly, the novel alignment loss and contrastive loss are constructed to supervise the student model to generate more satisfactory features for the CPC. 
Extensive experiments consist of comparison, ablation study, and qualitative visualization, which sufficiently illustrate the effectiveness and superiority of our approach ($\ie, 5.6\%$ accuracy improvement beyond the previous state-of-the-art approach on average). 

\section{Acknowledgement}
The work was supported in part by Key Area R\&D Program of Guangdong Province with grant No.2018B030338001, by the National Key R\&D Program of China with grant No.2018YFB1800800, by Shenzhen Outstanding Talents Training Fund, by Guangdong Research Project No.2017ZT07X152, by NSFC-Youth 61902335, by Guangdong Regional Joint Fund-Key Projects 2019B1515120039, by The National Natural Science Foundation Fund of China (61931024), by helixon biotechnology company Fund and CCF-Tencent Open Fund.

\bibliographystyle{splncs04}
\bibliography{mybibliography}
\end{document}